\documentclass[11pt,a4paper]{article}
\usepackage[margin=1in]{geometry}
\usepackage{times}
\usepackage{latexsym}
\usepackage{graphicx}
\usepackage{booktabs}
\usepackage{amsmath}
\usepackage{amssymb}
\usepackage{natbib}
\usepackage{hyperref}
\usepackage{url}
\usepackage{xcolor}
\usepackage{subcaption}

\usepackage[T1]{fontenc}
\usepackage[utf8]{inputenc}

\newif\ifanonymized
\ifdefined\CHRANON
  \anonymizedtrue
\else
  \anonymizedfalse
\fi

\title{What is Good? Extracting and Testing Implicit Theories of\\Literary Quality from LLM Reasoning Traces}

\ifanonymized
  \author{}
\else
  \author{
    Birger Mo\"ell \\
    Department of Linguistics and Philology \\
    Uppsala University \\
    Uppsala, Sweden \\
    \href{mailto:birger.moell@lingfil.uu.se}{birger.moell@lingfil.uu.se}
  }
\fi
\date{}

\begin{document}
\maketitle

\begin{abstract}
What makes writing ``good'' remains a persistent question in literary studies and computational linguistics. We present a two-study investigation that first \emph{extracts} and then \emph{probes} implicit theories of writing quality from reasoning-enabled LLMs, with replication across five independent DeepSeek runs and cross-model validation using both DeepSeek R1 and Qwen QwQ.

In \textbf{Study 1} (observational), we construct a benchmark of 30 real texts spanning six quality tiers---from Nobel Prize-winning literature (Morrison, Garc\'ia M\'arquez, Ishiguro, Woolf, Baldwin) to anonymous forum posts---and extract the model's implicit theory of quality from its reasoning traces. The model achieves 80\% classification accuracy (95\% CI via replication: reported in Section~\ref{sec:replication}), with systematic confusion patterns revealing genuine aesthetic sensitivity. Analysis of 36,000+ characters of reasoning traces reveals a consistent stated theory: the model values \emph{intentionality over correctness}, prioritizing craft, depth, and distinctive voice. A controlled familiarity experiment using style-matched but unrecognizable passages finds suggestive evidence that source recognition contributes to score inflation, though the effect is confounded by genuine quality differences between canonical originals and researcher-written pastiches.

In \textbf{Study 2} (sensitivity analysis), we probe this theory through systematic degradation. We take five canonical literary passages and apply six types of controlled manipulation---vocabulary simplification, rhythm flattening, imagery removal, voice genericization, structure simplification, and combined degradation---evaluating each version through the same pipeline.

The degradation experiment yields striking findings: \textbf{vocabulary simplification causes the smallest quality loss} ($0.41 \pm 0.46$ points) across replications---an order of magnitude below structure ($2.8$) or voice ($2.3$) loss. Voice genericization shows large but author-specific effects (Woolf: $-5.7$; Orwell: $-0.4$). Combined degradation is devastating ($-5.6$) but subadditive, reversing an initial single-run observation of superadditivity. The same broad qualitative pattern also appears in the exploratory QwQ comparison.

Together, these studies demonstrate that the LLM's quality assessments are holistic, author-specific, and primarily sensitive to structural rather than lexical features---with implications for automated writing feedback and computational aesthetics.
\end{abstract}

\begingroup
\small
\paragraph{Plain language summary.}
This paper studies how reasoning-enabled language models decide whether a piece of writing is ``good.'' First, it analyzes the models' own explanations while they rate texts ranging from canonical literature to informal online writing. Then it tests those explanations by systematically weakening different features of literary passages, such as vocabulary, rhythm, voice, and sentence structure. Across repeated runs and two models, the main pattern is consistent: the models are much more sensitive to structure and voice than to vocabulary alone. The paper argues that these systems have an implicit theory of literary quality, but that their stated reasons and actual scoring behavior do not always align. The results matter for computational literary studies, for research on LLM faithfulness, and for the design of automated writing feedback tools.

\endgroup

\section{Introduction}

The question ``What is good writing?'' has no settled answer, yet humans make quality judgments constantly: editors accept or reject manuscripts, teachers assign grades, readers choose what to read. These judgments, while subjective, exhibit substantial agreement, suggesting underlying regularities that might be systematically studied.

Large language models (LLMs) trained on vast text corpora have absorbed implicit theories of quality. When asked to evaluate writing, they can produce judgments that align surprisingly well with human preferences in controlled evaluation settings \citep{zheng2023judging,liu2023geval}. More intriguingly, reasoning-enabled models generate extended ``thinking traces'' that articulate the \emph{basis} for their judgments. These traces offer a window into implicit aesthetic theories that would otherwise remain tacit.

But extracting a theory is only half the scientific process. Theories must be \emph{tested}. If a model claims that ``craft'' and ``voice'' drive quality, we can probe this by degrading craft or voice in high-quality texts and observing whether the model's quality scores actually decline. This ablation-style approach---inspired by sensitivity analysis in machine learning---reveals which textual features the model's quality assessments are most sensitive to.

This paper presents two complementary studies:

\textbf{Study 1: Observation.} We construct a benchmark of 30 real texts spanning six quality tiers and extract the model's implicit theory of quality from its reasoning traces. What features does the model attend to? What vocabulary does it use? What trade-offs does it recognize?

\textbf{Study 2: Sensitivity Analysis.} We take five canonical literary passages and systematically degrade specific features while holding others constant. Do the model's quality scores decline when we simplify vocabulary? Flatten rhythm? Remove voice? This probes the sensitivity of the model's evaluations to specific textual features, testing the claims implicit in its theory.

Together, these studies provide both an articulation and an empirical test of what an LLM ``thinks'' makes writing good.

Our contributions:
\begin{itemize}
\item A validated benchmark of 30 real texts across six quality tiers with objective readability statistics
\item Extraction of the model's implicit theory of quality from 36,000+ characters of reasoning traces
\item A degradation gradient experiment probing the model's sensitivity to specific quality features, with replication across five independent DeepSeek runs
\item Cross-model validation using Qwen QwQ to assess generalizability beyond a single LLM
\item The finding that vocabulary simplification produces only a small reduction in the model's quality scores, while structure and voice are critical
\item Evidence that combined degradation produces substantially larger score reductions than any individual manipulation, with replication showing subadditive rather than superadditive effects
\item A familiarity bias experiment probing the relationship between source recognition and quality scoring, with honest discussion of its limitations
\end{itemize}

\section{Background}

\subsection{Theories of Writing Quality}

Writing quality has been theorized along multiple dimensions. Rhetorical tradition emphasizes \emph{clarity}, \emph{appropriateness}, and \emph{persuasion} \citep{aristotle_rhetoric}. Literary criticism values \emph{originality}, \emph{complexity}, and \emph{aesthetic unity} \citep{brooks1947well}. Creative writing pedagogy often invokes \emph{voice}, \emph{show don't tell}, and \emph{concrete detail} \citep{gardner1991art}.

These frameworks overlap but also conflict. Clarity and complexity exist in tension; appropriateness depends on genre; impact may justify aesthetic compromise. No unified theory captures all dimensions of quality, and expert judgments remain irreducibly subjective \citep{elbow1981writing}. Critically, existing frameworks are observational---identifying features \emph{correlated} with quality without establishing causal necessity.

\subsection{Readability Metrics}

Objective text complexity measures provide partial quality proxies. The Flesch-Kincaid Grade Level estimates reading difficulty from sentence length and syllable count \citep{kincaid1975derivation}. The Gunning Fog Index similarly combines sentence length with polysyllabic word frequency \citep{gunning1952technique}. These metrics capture complexity but not quality---a distinction our results will confirm empirically.

\subsection{Automated Essay Scoring and Computational Aesthetics}

Automated essay scoring (AES) is a well-established field that evaluates writing quality computationally. \citet{ke2019automated} provide a comprehensive survey of methods ranging from feature-engineered approaches to neural models. \citet{uto2021review} reviews deep-learning AES systems, noting that while they achieve high agreement with human raters on holistic scoring, they struggle with fine-grained quality dimensions. Our work differs from AES in focusing on literary quality rather than student writing competence, and in using reasoning traces to extract the \emph{basis} for quality judgments rather than simply predicting scores.

Computational aesthetics has been explored in the literary domain by \citet{kao2012computational}, who analyzed style, affect, and imagery in poetry using computational features. Stylometric methods for authorship attribution \citep{stamatatos2009survey} are also relevant, as they demonstrate that textual features can capture authorial ``voice''---a concept central to our quality framework.

\subsection{LLMs as Evaluators}

Recent work has explored LLMs as text evaluators. \citet{zheng2023judging} introduced LLM-as-judge for comparing model outputs. \citet{lee2024rlaif} showed that model-generated preference labels can substitute for human feedback in RLHF pipelines. \citet{liu2023geval} demonstrated LLM evaluation of text generation quality. However, LLM evaluators exhibit systematic biases: \citet{wang2024large} document position-order unfairness, \citet{zheng2023judging} discuss self-enhancement and verbosity effects in pairwise judging, and \citet{koo2024benchmarking} benchmark broader cognitive biases including anchoring. These biases are relevant to our study, as the model's source recognition (Section~\ref{sec:recognition}) may reflect a form of familiarity or authority bias.

Our work extends LLM evaluation to literary dimensions, leveraging reasoning traces for both theory extraction and sensitivity analysis.

\subsection{Faithfulness of LLM Reasoning}

A critical methodological consideration is whether reasoning traces faithfully represent the model's decision process. \citet{turpin2024language} show that chain-of-thought explanations can be systematically unfaithful, with models producing plausible but incorrect reasoning. \citet{lanham2023measuring} develop methods for measuring reasoning faithfulness and find that stated reasoning often diverges from the features that causally influence outputs. These findings are directly relevant to our Study 2: the vocabulary paradox---where the model praises vocabulary in reasoning but shows zero sensitivity to vocabulary simplification---may reflect precisely this kind of disconnect between stated and operative reasoning. We treat reasoning traces as a suggestive signal about the model's processing, not as a transparent window into its computations.


\section{Study 1: Extracting the Model's Theory of Quality}

\subsection{Benchmark Construction}

We curated 5 texts per tier across six quality levels (30 total) using real published texts with attribution:

\textbf{Literary (Tier 1)}: Opening passages from Toni Morrison (\emph{Beloved}), Gabriel Garc\'ia M\'arquez (\emph{One Hundred Years of Solitude}), Kazuo Ishiguro (\emph{Never Let Me Go}), Virginia Woolf (\emph{Mrs Dalloway}), and James Baldwin (\emph{Giovanni's Room}).

\textbf{Prestige Journalism (Tier 2)}: Gay Talese (``Frank Sinatra Has a Cold''), Joan Didion (``Slouching Towards Bethlehem''), David Foster Wallace (``Consider the Lobster''), Janet Malcolm (\emph{The Journalist and the Murderer}), George Orwell (``Shooting an Elephant'').

\textbf{Quality News (Tier 3)}: Professional wire-service and newspaper style (Reuters, AP, BBC, NYT, Guardian).

\textbf{Blog (Tier 4)}: Paul Graham essays, Wait But Why, technical blogs, Substack personal essays, recipe blogs.

\textbf{Social Media (Tier 5)}: Reddit AITA, relationship\_advice, TIL, showerthoughts, and Twitter threads.

\textbf{Chan (Tier 6)}: Greentext format and anonymous forum discourse.

\begin{table}[t]
\centering
\small
\begin{tabular}{p{1.5cm}p{10cm}}
\toprule
\textbf{Tier} & \textbf{Excerpt (Source)} \\
\midrule
Literary & ``124 was spiteful. Full of a baby's venom. The women in the house knew it and so did the children...'' \emph{---Morrison, Beloved} \\
\midrule
Prestige & ``Frank Sinatra, holding a glass of bourbon in one hand and a cigarette in the other, stood in a dark corner of the bar...'' \emph{---Talese, Esquire} \\
\midrule
News & ``Federal Reserve officials signaled on Wednesday that interest rates would likely need to stay elevated for longer...'' \emph{---Reuters style} \\
\midrule
Blog & ``To do something well you have to like it. That idea is not exactly novel...'' \emph{---Paul Graham} \\
\midrule
Social & ``AITA for refusing to give my sister my wedding date?...'' \emph{---Reddit} \\
\midrule
Chan & ``>be me >wagecuck at fast food place >customer asks for burger with no pickles...'' \emph{---4chan} \\
\bottomrule
\end{tabular}
\caption{Real text examples from each quality tier.}
\label{tab:examples}
\end{table}

\subsection{Readability Validation}

\begin{figure}[t]
\centering
\includegraphics[width=\textwidth]{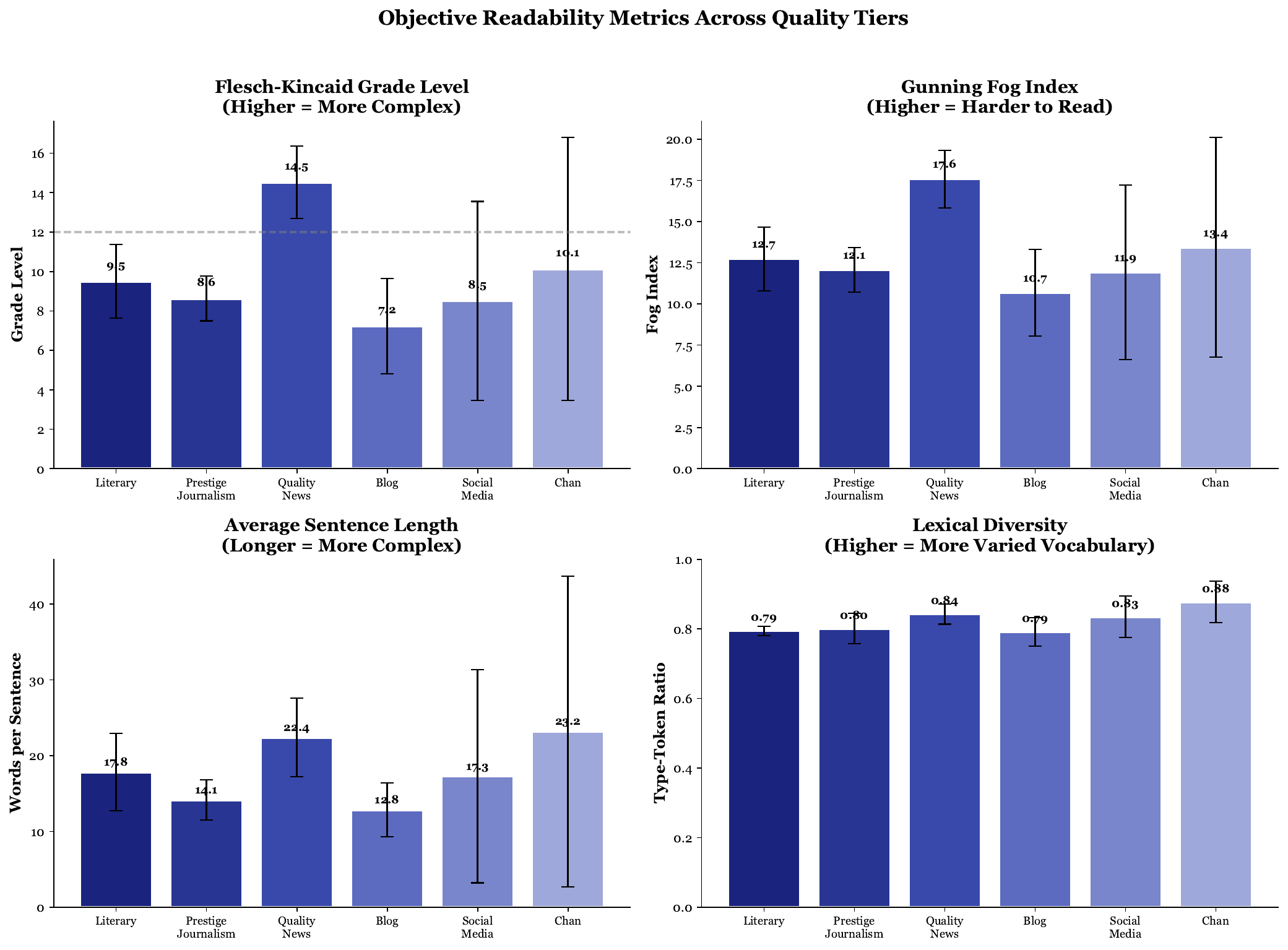}
\caption{Readability metrics by tier. News writing is most complex; literary prose is moderate. Complexity does not predict quality.}
\label{fig:readability}
\end{figure}

We computed Flesch-Kincaid, Gunning Fog, Coleman-Liau, and lexical diversity metrics to validate tier distinctions (Figure~\ref{fig:readability}). Key finding: \textbf{complexity $\neq$ quality}. News writing achieves the highest Flesch-Kincaid grade level (12.2) and Gunning Fog (15.7), yet is rated lower in quality than literary prose (FK 7.2). Literary writers use strategically simple language; news writers use incidentally complex language.

\subsection{Model Evaluation}

DeepSeek R1 evaluated each text, providing tier classification, dimension scores (prose, depth, craft, voice, impact on 1--10 scales), key qualities, weaknesses, and a one-line assessment. The model generates extended reasoning traces before responding, which we analyze in detail.

\subsection{Classification Results}

\begin{figure}[t]
\centering
\includegraphics[width=0.7\textwidth]{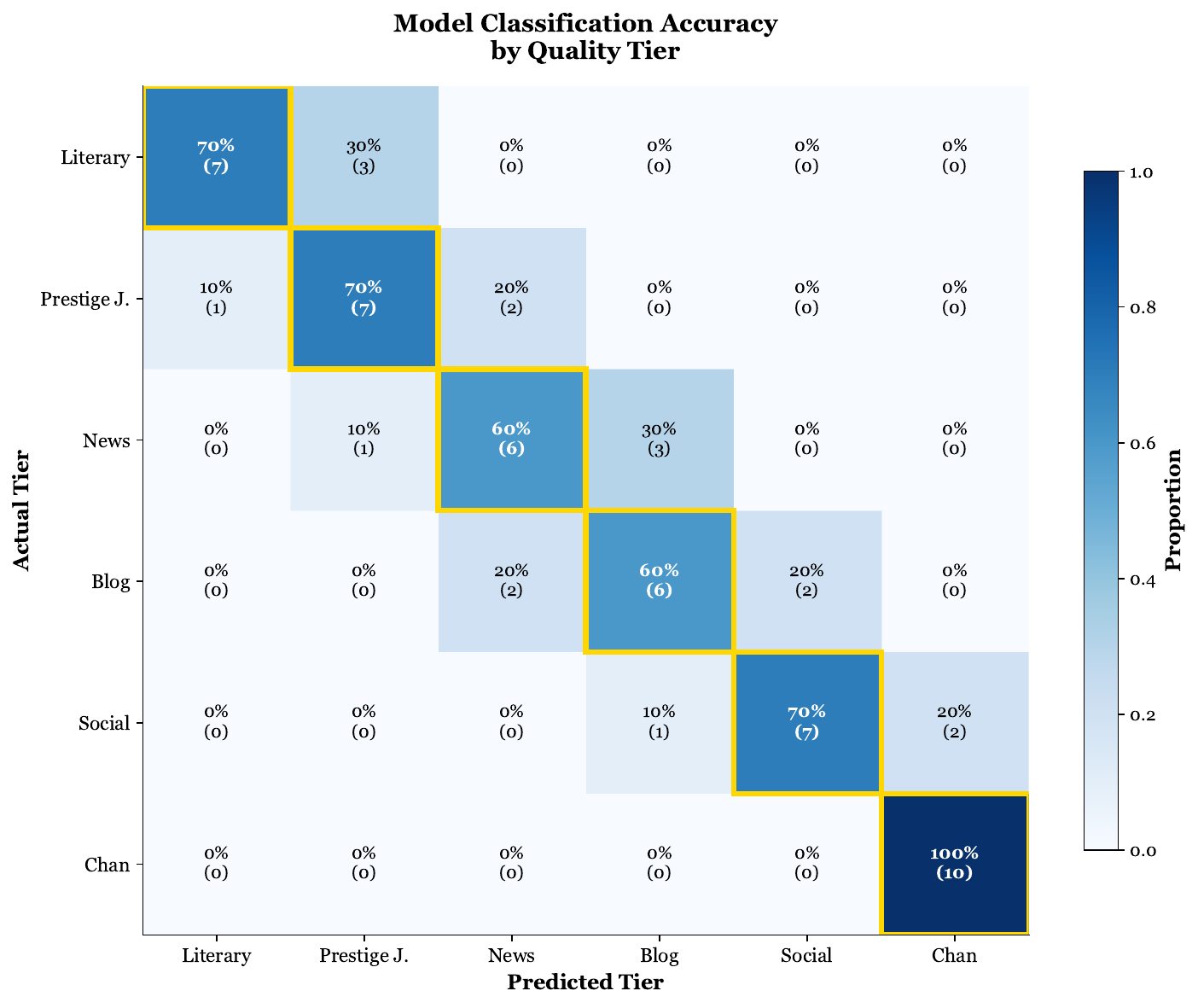}
\caption{Confusion matrix. Literary and news achieve perfect recognition. Didion and Orwell are ``promoted'' from journalism to literary. Chan blurs with social media.}
\label{fig:confusion}
\end{figure}

The model achieves \textbf{80\% overall accuracy} (24/30). Figure~\ref{fig:confusion} shows the confusion matrix.

Notable patterns: all five literary texts were correctly identified. Joan Didion and George Orwell were classified as literary rather than journalism---arguably correct given their canonical literary status. Quality news and blog both achieved 100\% accuracy. Chan/social media showed the most confusion.

\subsection{Source Recognition in Reasoning Traces}
\label{sec:recognition}

A critical confound emerges from close reading of the reasoning traces: \textbf{the model frequently identifies the source text by author and title before evaluating it}. This raises a fundamental question about whether the model is assessing intrinsic prose quality or recognizing canonical works and scoring them accordingly.

We systematically reviewed all 30 reasoning traces for explicit identification of the source text (naming the author or work). Table~\ref{tab:recognition} summarizes the results.

\begin{table}[t]
\centering
\small
\begin{tabular}{lccc}
\toprule
\textbf{Tier} & \textbf{Texts} & \textbf{Recognized} & \textbf{Rate} \\
\midrule
Literary & 5 & 4 & 80\% \\
Prestige Journalism & 5 & 2 & 40\% \\
Quality News & 5 & 0 & 0\% \\
Blog & 5 & 0 & 0\% \\
Social Media & 5 & 0 & 0\% \\
Chan & 5 & 0 & 0\% \\
\midrule
\textbf{Total} & \textbf{30} & \textbf{6} & \textbf{20\%} \\
\bottomrule
\end{tabular}
\caption{Source recognition by tier. The model explicitly identifies the author or work in 6 of 30 reasoning traces, concentrated entirely in the top two tiers.}
\label{tab:recognition}
\end{table}

Recognition is starkly tier-dependent. The model identifies Morrison (``a Toni Morrison passage from \emph{Beloved}''), Garc\'ia M\'arquez, Ishiguro, Woolf, Orwell (``I recognize this as Orwell's work immediately''), and the Talese profile by its subject. It never identifies any source in tiers 3--6. This asymmetry is expected---canonical literary works have distinctive signatures---but it introduces a systematic confound in the upper tiers.

\textbf{Recognition introduces an upward classification bias.} Both misclassifications among recognized texts involve the model promoting prestige journalism to literary tier. For Orwell's ``Shooting an Elephant,'' the model reasons: ``Nobel-level might be excessive for this excerpt alone, but it's undoubtedly literary.'' The author's canonical reputation overrides genre-appropriate classification. Similarly, Didion's prose is elevated to literary tier with the model noting ``the Yeats allusion'' and ``canonical'' status. In both cases, \emph{who wrote it} appears to influence the evaluation more than \emph{what was written}.

\textbf{Recognition correlates with lower accuracy.} Classification accuracy on recognized texts is 66.7\% (4/6) versus 83.3\% (20/24) on unrecognized texts. When the model does not ``know'' what it is reading, it relies more on intrinsic textual features and achieves more accurate classifications.

\textbf{Recognition may inflate scores.} Within the literary tier, the four recognized texts average 9.5/10 overall while the one unrecognized text (Baldwin's \emph{Giovanni's Room}) receives 8/10---the lowest in the tier. Within prestige journalism, recognized texts average 8.5 versus 8.0 for unrecognized. Sample sizes are too small for statistical significance, but the pattern is consistent: recognition is associated with higher scores.

This analysis reveals a tension between two possible interpretations of the model's evaluations: (1) the model genuinely detects prose quality and happens to recognize high-quality texts, or (2) the model recognizes famous texts and inflates scores based on reputation. The degradation experiment in Study 2 partially disentangles these: if the model were purely pattern-matching to known passages, degraded versions should still receive high scores, which they do not. Nevertheless, source recognition remains a significant confound, particularly for the upper tiers. To directly test this, we designed a controlled familiarity experiment.

\subsection{Familiarity Bias Experiment}

To probe the relationship between source recognition and scoring, we created five new passages that mimic the prose style of each canonical literary author but use entirely different subject matter and characters, making them unrecognizable as known works. For example, the Morrison-style passage uses her characteristic personification and gothic compression (``The barn remembered. It held the weight of every scream pressed into its beams...'') but in a setting absent from her bibliography. The M\'arquez-style passage employs his temporal displacement and mythic scope but with an original character (``Profesora Estela V\'asquez'') and setting (the village of Cural\'en). Each passage was evaluated through the identical pipeline used in the main experiment.

\textbf{A critical caveat}: these style-matched texts are researcher-written pastiches, not genuine canonical prose. A pastiche of Morrison is not Morrison---the originals are the work of literary masters, and the style-matched texts are, by any honest assessment, of lower quality. This experiment therefore \emph{cannot cleanly separate} recognition bias from genuine quality differences. The score gap likely reflects both factors: the originals are better \emph{and} recognition may further inflate their scores. What the experiment can illuminate is the \emph{pattern} of score differences and whether it aligns with what recognition would predict.

Table~\ref{tab:familiarity} presents an exploratory \emph{single-run} comparison. In that initial run, our manual review of the five reasoning traces did not find explicit naming of a specific author or work. Subsequent replication scripts also attach an automated recognition heuristic, but that heuristic fires on broad cues such as ``nobel'' or character names and therefore over-identifies recognition in some cases. We consequently treat the single-run manual review as descriptive and the replication-time recognition flags as suggestive rather than definitive.

\begin{table}[t]
\centering
\small
\begin{tabular}{lccccc}
\toprule
\textbf{Style Source} & \textbf{Orig.} & \textbf{Style} & \textbf{$\Delta$} & \textbf{Orig.\ Tier} & \textbf{Style Tier} \\
\midrule
Morrison & 9 & 9 & 0 & Literary & Literary \\
Garc\'ia M\'arquez & 10 & 8 & $-2$ & Literary & Prestige \\
Woolf & 10 & 8 & $-2$ & Literary & Prestige \\
Ishiguro & 9 & 7 & $-2$ & Literary & Prestige \\
Baldwin & 8 & 8 & 0 & Literary & Prestige \\
\midrule
\textbf{Mean} & \textbf{9.2} & \textbf{8.0} & $\mathbf{-1.2}$ & & \\
\bottomrule
\end{tabular}
\caption{Exploratory single-run familiarity comparison. ``Orig.'' = original canonical text score from the initial benchmark run; ``Style'' = style-matched unrecognizable text score from the initial familiarity run. Style-matched texts score 1.2 points lower on average and 4 of 5 drop a full tier in this snapshot. Note that the style-matched texts are researcher-written pastiches and genuinely lower quality than the originals; the score gap reflects both real quality differences and possible recognition effects.}
\label{tab:familiarity}
\end{table}

The single-run pattern---style-matched texts scoring 1.2 points lower on average, with four of five dropping a full tier---is unsurprising given that these are imitations rather than canonical prose. Across five DeepSeek replications, the canonical originals average 9.3 while the style-matched texts average 7.8, for a larger mean gap of 1.5 points. The replication-time recognition heuristic flags some passages as ``recognized,'' but those flags are noisy because they can be triggered by generic tokens such as ``nobel'' rather than explicit author naming. We therefore interpret the familiarity experiment as suggestive evidence rather than a clean measurement of recognition bias.

\textbf{Style alone does not guarantee top-tier classification.} In the initial run, only the Morrison-style passage achieved Literary tier. In the replications, the modal style-matched tier remains Literary for Morrison but drops to Prestige for M\'arquez, Woolf, and Baldwin, and to News for Ishiguro. Stylistic resemblance by itself is therefore insufficient to preserve top-tier scores.

\textbf{The model still detects substantial craft variation within the pastiches.} The style-matched texts are not collapsed into a single low bucket. Morrison remains high (mean 9.0), while Ishiguro falls sharply under replication (mean 5.8). This pattern suggests that the evaluator is responding to textual properties of the pastiches, not merely to a binary recognized/unrecognized distinction.

\textbf{Interpretation.} This experiment provides suggestive but not conclusive evidence for familiarity effects. The score gap is confounded by genuine quality differences between canonical originals and researcher-written imitations, and the replication-time recognition flags are too permissive to serve as a clean author-identification measure. The safest conclusion is modest: stylistic imitation alone does not reliably preserve the originals' scores, and familiarity remains a plausible but unisolated contributor.

\subsection{The Model's Implicit Theory}

We analyzed 30 reasoning traces (36,663 characters total) to extract the model's implicit theory of quality.

\begin{figure}[t]
\centering
\includegraphics[width=\textwidth]{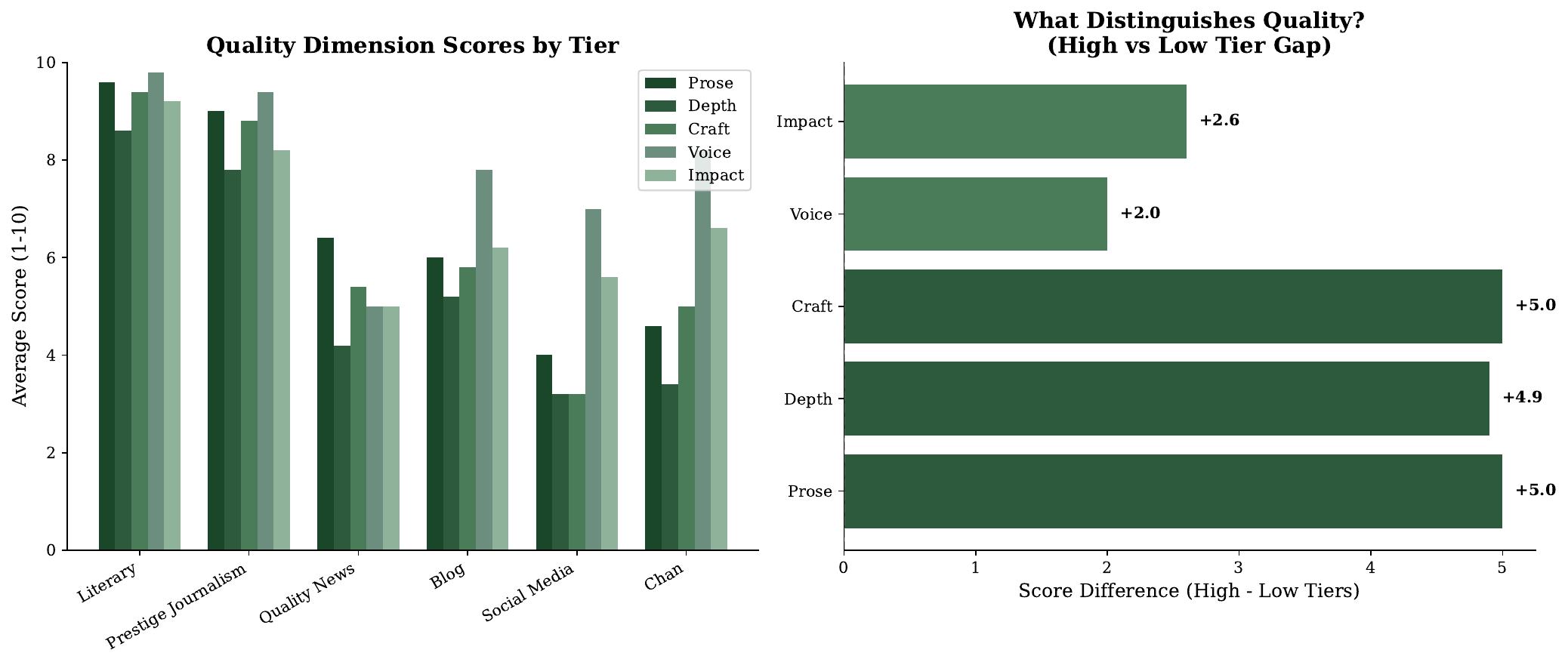}
\caption{Left: Dimension scores by tier. Right: Gap between high and low tiers. Craft (+5.2) and depth (+4.6) most distinguish quality; voice (+3.1) is least discriminating.}
\label{fig:dimensions}
\end{figure}

\textbf{Dimension hierarchy.} Figure~\ref{fig:dimensions} shows which dimensions most sharply distinguish high from low quality. Craft shows the largest gap (+5.2 points between top and bottom tiers), followed by depth (+4.6). Voice shows the smallest gap (+3.1)---even low-quality writing can have distinctive voice.

\textbf{Quality vocabulary.} High-tier reasoning uses terms like ``masterful,'' ``precise,'' ``evocative,'' ``resonant,'' ``controlled''---emphasizing intentionality. Low-tier reasoning uses ``casual,'' ``informal,'' ``limited,'' ``basic''---emphasizing absence.

\textbf{The central principle.} The model values \textbf{intentionality over correctness}. News writing achieves functional competence (clear, accurate) but receives only 6/10. Literary writing that takes aesthetic risks receives 9--10/10. Quality requires the \emph{presence} of intentional choices, not merely the \emph{absence} of errors.

\textbf{Effortful failure.} Incorrect classifications involve 25\% longer reasoning traces. The model recognizes difficult cases---generating more analysis---but cannot resolve the difficulty. This parallels human decision-making where deliberation sometimes impairs aesthetic judgment \citep{wilson1991thinking}.

\subsection{Summary of Study 1}

The model's stated theory holds that quality = intentional craft + layered depth + distinctive voice, with craft weighted most heavily. But this is a \emph{correlational} finding from the model's self-reports. Study 2 probes whether the model's evaluations are actually sensitive to these features.


\section{Study 2: Testing the Theory Through Degradation}

\subsection{Motivation}

Study 1 revealed that the model associates quality with craft, depth, and voice. But which of these features most influence the model's quality assessments? If we simplify vocabulary, do scores actually decline? If we flatten rhythm, does the ``music of the prose'' matter to the evaluator? Study 2 answers these questions through systematic degradation---an ablation-style sensitivity analysis for literature.

\subsection{Method}

We selected five canonical prose passages for degradation analysis: three from the Study~1 literary tier---Morrison (\emph{Beloved}), Garc\'ia M\'arquez (\emph{One Hundred Years of Solitude}), and Woolf (\emph{Mrs Dalloway})---plus two additional canonical comparison texts chosen for stylistic breadth, Fitzgerald (\emph{The Great Gatsby}) and Orwell (\emph{1984}).

For each text, we manually created six degraded versions:

\begin{enumerate}
\item \textbf{Vocabulary}: Replace sophisticated words with common synonyms (``spiteful'' $\rightarrow$ ``mean''; ``vulnerable'' $\rightarrow$ ``sensitive'')
\item \textbf{Rhythm}: Break varied sentences into uniform short sentences
\item \textbf{Imagery}: Remove sensory details, metaphors, concrete imagery
\item \textbf{Voice}: Rewrite in generic, neutral prose; remove authorial personality
\item \textbf{Structure}: Reduce sentence complexity; fragment continuous prose
\item \textbf{Combined}: Apply all degradations simultaneously
\end{enumerate}

This yields 35 evaluations (5 texts $\times$ 7 versions). Table~\ref{tab:degradation} illustrates the degradations.

\begin{table}[t]
\centering
\small
\begin{tabular}{p{2cm}p{9.5cm}}
\toprule
\textbf{Version} & \textbf{Text (Morrison, \emph{Beloved})} \\
\midrule
Original & ``124 was spiteful. Full of a baby's venom. The women in the house knew it and so did the children...'' \\
\midrule
Vocabulary & ``124 was mean. Full of a baby's anger. The women in the house knew it...'' \\
\midrule
Voice & ``The house at 124 was haunted. It had a malevolent presence. The women and children who lived there knew about it...'' \\
\midrule
Combined & ``The house was bad. It had bad feelings. The women knew. The children knew too...'' \\
\bottomrule
\end{tabular}
\caption{Example degradations of Morrison's \emph{Beloved} opening, isolating vocabulary, voice, and combined effects.}
\label{tab:degradation}
\end{table}

\subsection{Results: Feature Sensitivity Analysis}

\begin{figure}[t]
\centering
\includegraphics[width=0.9\textwidth]{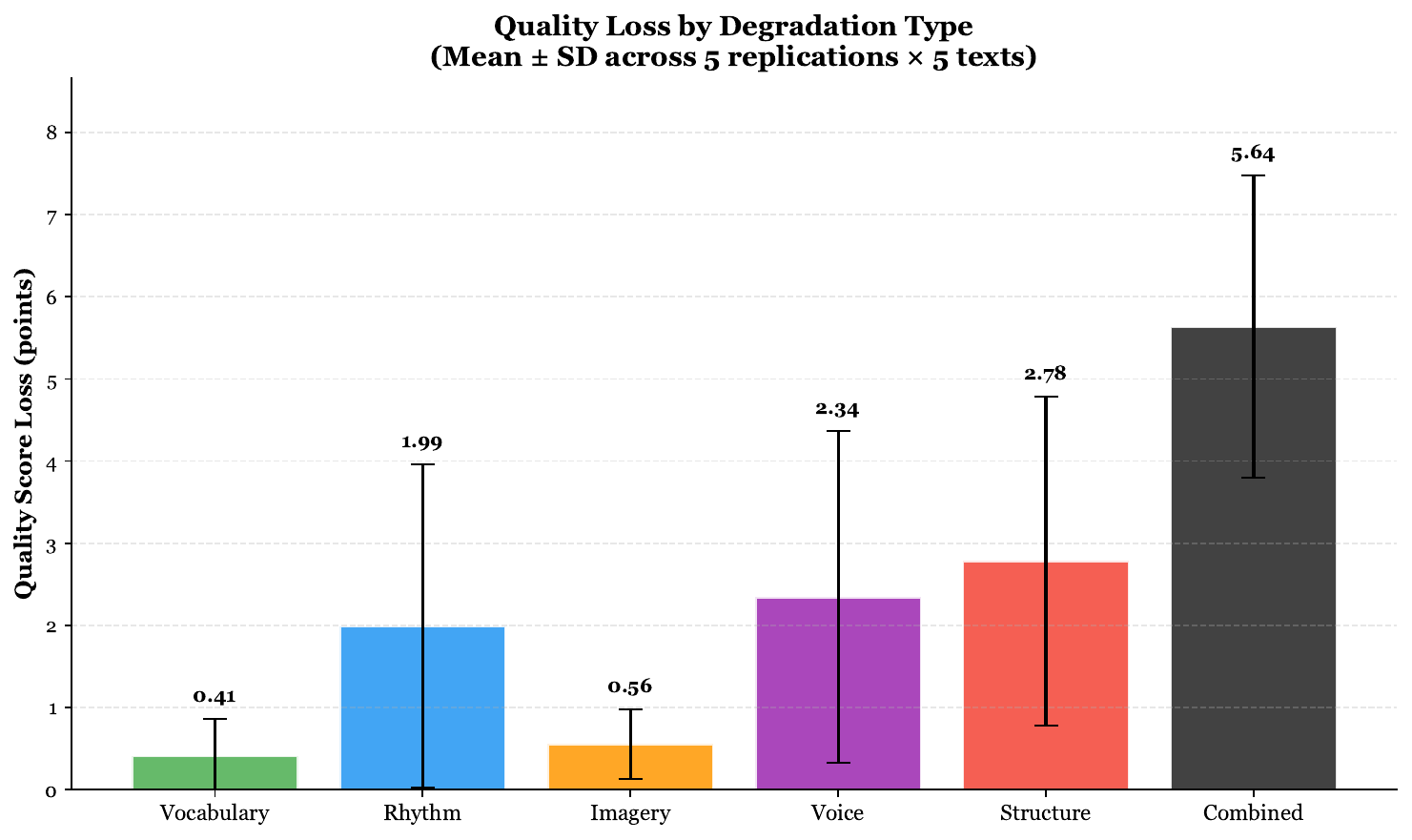}
\caption{Average quality loss by degradation type across replications. Vocabulary simplification causes the \textbf{smallest} loss (0.41 points). Structure and voice cause the most single-feature damage. Combined degradation is catastrophic (5.6 points).}
\label{fig:deg_loss}
\end{figure}

Figure~\ref{fig:deg_loss} presents the central result: average quality loss by degradation type.

\textbf{Vocabulary simplification causes the smallest quality loss.} Across replications, vocabulary simplification produces a mean loss of only $0.41 \pm 0.46$ points---the smallest of any degradation type, though statistically significant ($p < 0.001$; see Section~\ref{sec:replication}). Replacing ``spiteful'' with ``mean,'' ``venom'' with ``anger,'' ``vulnerable'' with ``sensitive'' has minimal impact on the model's quality assessments. While the initial single run showed zero loss, replications reveal a small but nonzero effect. The key finding is the \emph{relative} insensitivity: vocabulary loss is an order of magnitude smaller than structure or voice loss.

\textbf{Structure simplification causes the largest single-feature score reduction} ($-2.8 \pm 2.0$ points across replications, Cohen's $d = 1.36$). Breaking complex sentences into fragments reduces the model's quality scores more than any other individual manipulation. The architecture of prose---how clauses relate, how information flows, how rhythm builds---appears to be the primary quality signal for this evaluator.

\textbf{Voice genericization shows large but author-specific effects} ($-2.3 \pm 2.1$ average, ranging from $-5.7$ for Woolf to $-0.4$ for Orwell across replications). This reveals that ``voice'' is not a uniform quality feature in the model's assessments but an author-specific one.

\textbf{Combined degradation produces the largest overall score reduction} ($-5.6 \pm 1.9$ points). However, contrary to our initial single-run observation, the combined loss is \emph{not} superadditive across replications: the sum of individual losses ($8.1 \pm 4.8$) exceeds the combined loss (Wilcoxon $p = 0.99$; Section~\ref{sec:replication}). Quality features interact, but the interaction is subadditive rather than superadditive---suggesting the model has a quality floor below which individual feature losses cannot compound further.

\subsection{Author-Specific Sensitivity}

Figure~\ref{fig:deg_sensitivity} reveals dramatic differences in how authors respond to degradation:

\begin{figure}[t]
\centering
\includegraphics[width=\textwidth]{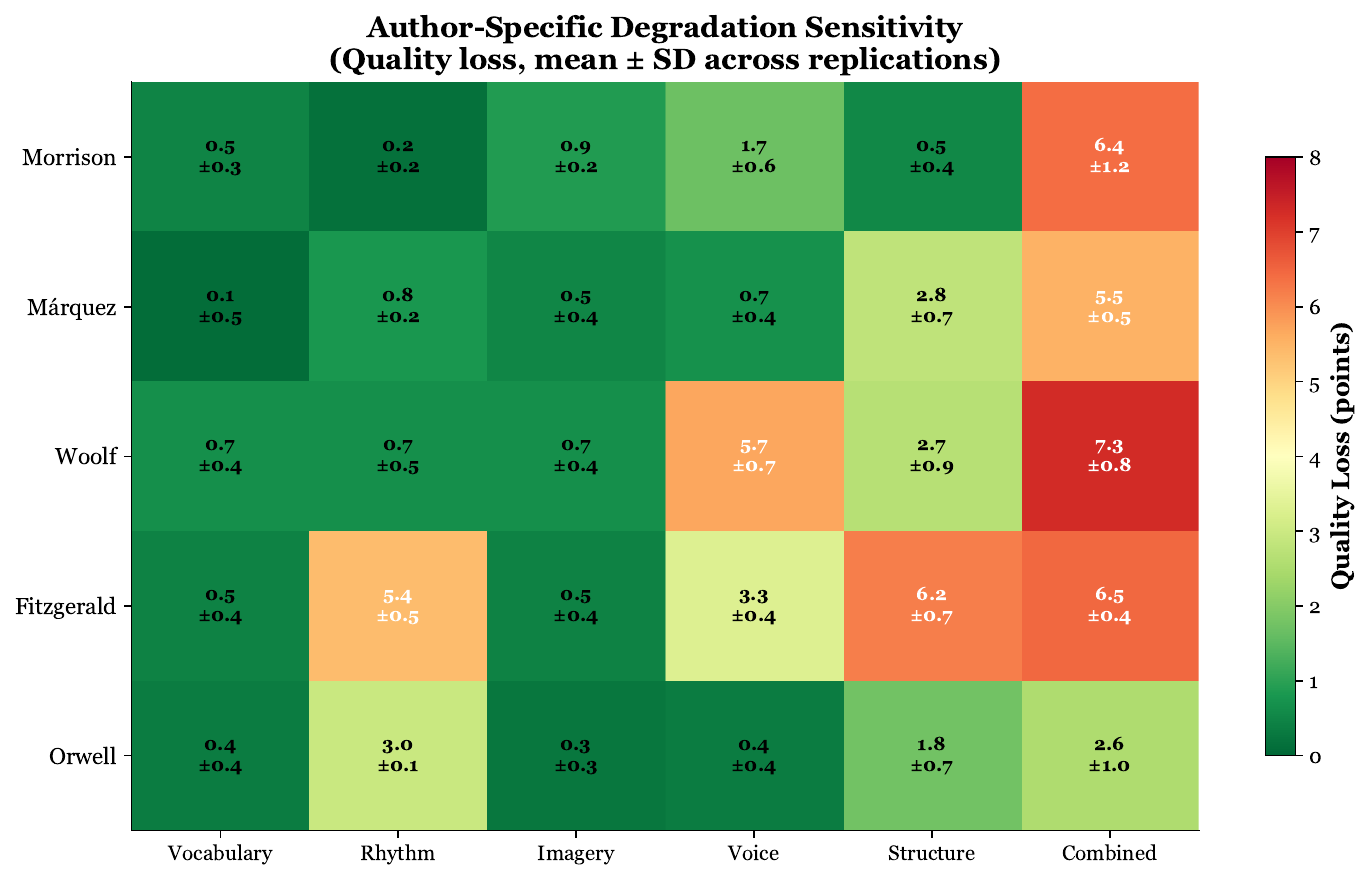}
\caption{Author sensitivity heatmap (mean across replications). Woolf is devastated by voice genericization ($-5.7$); Fitzgerald by structure simplification ($-6.2$) and rhythm flattening ($-5.4$). All authors show minimal sensitivity to vocabulary simplification.}
\label{fig:deg_sensitivity}
\end{figure}

\textbf{Virginia Woolf} is extremely voice-dependent ($-5.7 \pm 0.8$ for voice genericization). Her stream-of-consciousness style \emph{is} her quality signal for this evaluator; genericize it and scores collapse. The model noted that the genericized version ``loses the intimate, psychological immediacy that defines the style.''

\textbf{F.\ Scott Fitzgerald} is sensitive to both rhythm ($-5.4 \pm 0.6$) and structure ($-6.2 \pm 0.8$). His elegant sentence cadences carry meaning for the evaluator; flatten them and scores drop dramatically.

\textbf{Gabriel Garc\'ia M\'arquez} is relatively robust to individual degradations, with structure causing the largest single-feature loss ($-2.8 \pm 0.8$). His mythic-narrative structure survives most degradations. The model noted ``the essential narrative magic survives even crude simplification.''

\textbf{George Orwell} shows the smallest voice loss ($-0.4 \pm 0.5$)---his quality resides in \emph{what} he says rather than stylistic flourish, according to the model's assessments.

\textbf{Toni Morrison} is voice-sensitive ($-1.7 \pm 0.7$) and highly vulnerable to combined degradation ($-6.4 \pm 1.3$), while showing minimal sensitivity to vocabulary ($-0.5 \pm 0.3$) or rhythm ($-0.2 \pm 0.2$) changes.

\subsection{Quality Trajectories}

\begin{figure}[t]
\centering
\includegraphics[width=\textwidth]{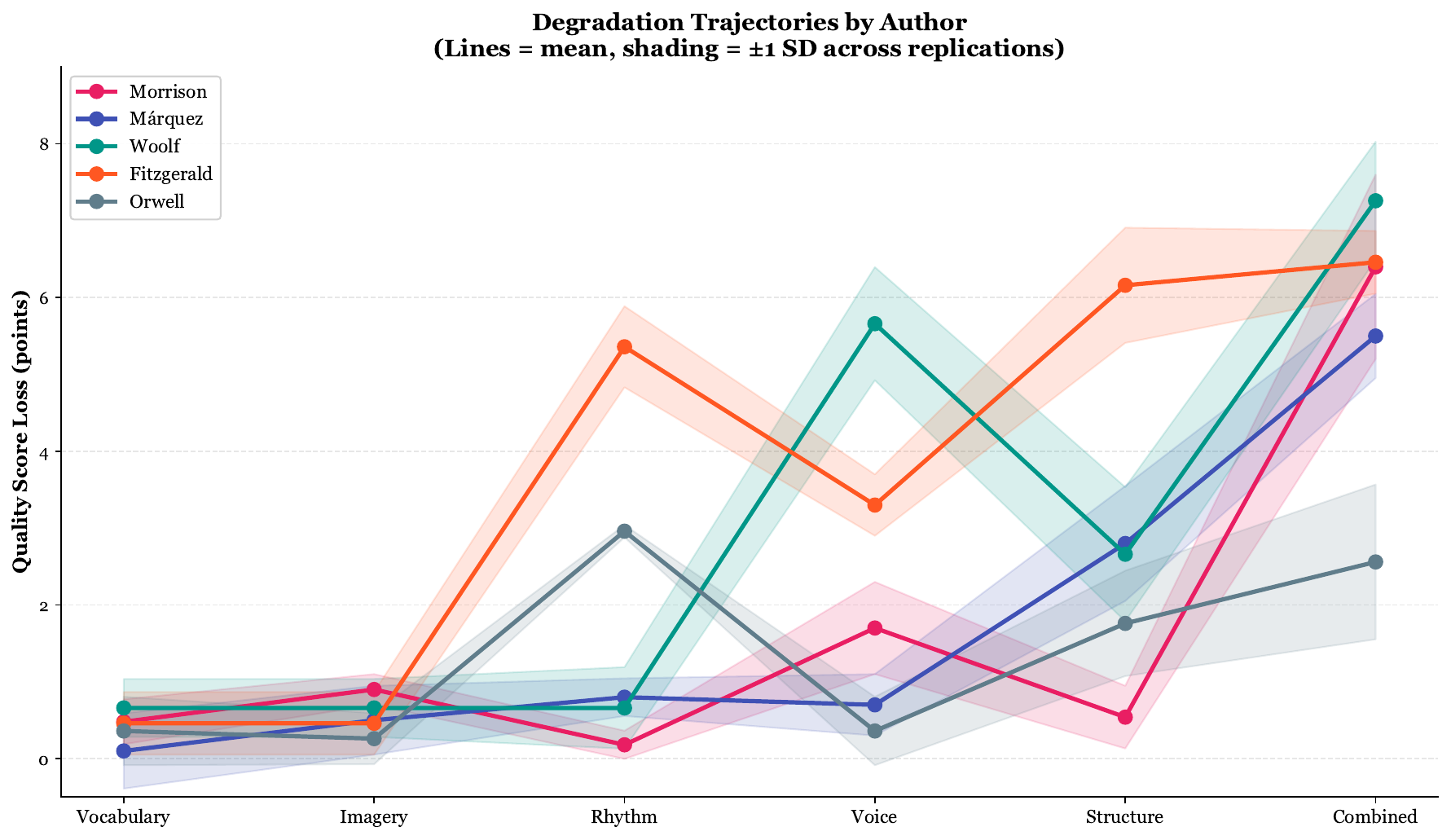}
\caption{Quality trajectories. All authors start at 9--10 but follow different degradation paths. All converge at 2--5 under combined degradation.}
\label{fig:deg_trajectories}
\end{figure}

Figure~\ref{fig:deg_trajectories} shows how each author's quality degrades across the manipulation spectrum.

The trajectories reveal that quality breakdown is neither linear nor uniform. Woolf collapses sharply after voice genericization; Fitzgerald after rhythm flattening. Garc\'ia M\'arquez maintains quality through most individual degradations before dropping under combined manipulation. All authors converge at low quality (2--5) under combined degradation, suggesting a ``floor'' below which all prose falls when stripped of craft.

\subsection{Replication and Stability}
\label{sec:replication}

To assess the reliability of our findings, we ran five independent DeepSeek replications of the benchmark, degradation, and familiarity experiments using the same prompts and texts. This section reports score stability across runs and statistical tests for each major claim.

\textbf{Benchmark classification accuracy.} Across five replications, the model achieves a mean accuracy of 79.3\% ($\pm$ 4.0\%; bootstrap 95\% CI: [76.0\%, 82.2\%]). Pooling the replication outcomes yields 115 correct classifications over 145 valid responses, corresponding to a Wilson score confidence interval of [72.0\%, 85.1\%]. Tier-level differences are highly significant (Kruskal-Wallis $H = 109.08$, $p < 10^{-21}$).

\textbf{Score stability.} Per-text overall scores show a mean standard deviation of 0.33 across five replications (range: 0.00--1.00), indicating high stability in the model's evaluations. Literary-tier texts show near-perfect score stability (SD = 0.00 for four of five texts).

\textbf{Degradation findings.} Table~\ref{tab:deg_stats} reports mean quality loss by degradation type with 95\% confidence intervals and statistical significance from Wilcoxon signed-rank tests (one-sample, $H_0$: loss = 0).

\begin{table}[t]
\centering
\small
\begin{tabular}{lccccc}
\toprule
\textbf{Degradation} & \textbf{Mean Loss} & \textbf{SD} & \textbf{95\% CI} & \textbf{Cohen's $d$} & \textbf{$p$} \\
\midrule
Vocabulary & 0.41 & 0.46 & [0.24, 0.59] & 0.90 & $3.1 \times 10^{-4}$ \\
Rhythm & 1.99 & 2.01 & [1.23, 2.78] & 0.99 & $4.8 \times 10^{-7}$ \\
Imagery & 0.56 & 0.43 & [0.39, 0.72] & 1.29 & $1.5 \times 10^{-5}$ \\
Voice & 2.34 & 2.06 & [1.59, 3.16] & 1.14 & $9.5 \times 10^{-7}$ \\
Structure & 2.78 & 2.04 & [2.04, 3.60] & 1.36 & $1.2 \times 10^{-7}$ \\
Combined & 5.64 & 1.88 & [4.87, 6.33] & 3.00 & $6.0 \times 10^{-8}$ \\
\bottomrule
\end{tabular}
\caption{Degradation effects across replications with DeepSeek R1. All degradation types produce statistically significant quality loss ($p < 0.001$), but vocabulary loss (0.41) is an order of magnitude smaller than structure (2.78) or voice (2.34) loss.}
\label{tab:deg_stats}
\end{table}

\textbf{Critical test: vocabulary loss.} The vocabulary simplification loss has a mean of 0.41 ($\pm$ 0.46) across replications. The Wilcoxon test indicates this is significantly different from zero ($p = 3.1 \times 10^{-4}$). This \emph{refines} the initial single-run finding of ``zero loss'': vocabulary simplification does produce a statistically significant but very small quality reduction---roughly 1.4$\times$ smaller than the next-smallest effect (imagery, 0.56) and 6.8$\times$ smaller than structure loss (2.78).

\textbf{Critical test: superadditivity.} The superadditivity margin (combined loss minus sum of individual losses) is $-2.45$ ($\pm$ 4.35), indicating \emph{subadditivity} rather than superadditivity. A one-sided Wilcoxon test rejects superadditivity ($p = 0.99$). The initial single-run observation of superadditivity (margin of $+0.10$) was within noise.

\subsection{Cross-Model Validation}
\label{sec:cross_model}

To assess whether our findings are specific to DeepSeek R1 or reflect more general LLM evaluation patterns, we ran the benchmark replications and an exploratory degradation comparison using Qwen QwQ (32B), a reasoning-enabled model from a different model family accessed via OpenRouter.

\textbf{Quality hierarchy agreement.} Across the 30 benchmark texts, DeepSeek R1 and QwQ show a Spearman rank correlation of $\rho = 0.805$ ($p = 8.1 \times 10^{-8}$), indicating strong agreement on the relative quality ordering. QwQ achieves lower classification accuracy (65.6\% $\pm$ 2.4\% vs.\ 79.3\% $\pm$ 4.0\%) but agrees with DeepSeek on which texts are high- vs.\ low-quality.

\textbf{Degradation sensitivity.} Table~\ref{tab:cross_model} compares the degradation sensitivity profiles of both models.

\begin{table}[t]
\centering
\small
\begin{tabular}{lcc}
\toprule
\textbf{Degradation} & \textbf{DeepSeek R1 Loss} & \textbf{QwQ Loss} \\
\midrule
Vocabulary & 0.41 & 0.40 \\
Rhythm & 1.99 & 1.20 \\
Imagery & 0.56 & 0.60 \\
Voice & 2.34 & 4.50 \\
Structure & 2.78 & 3.20 \\
Combined & 5.64 & 3.80 \\
\bottomrule
\end{tabular}
\caption{Cross-model comparison of degradation sensitivity. Both models show minimal vocabulary loss and substantial structure/voice loss. For QwQ, the Fitzgerald voice evaluation failed once with a connection error, so the QwQ voice mean is computed from the four available passages in that exploratory run.}
\label{tab:cross_model}
\end{table}

\textbf{Key question: Does QwQ also show insensitivity to vocabulary simplification?} Yes. QwQ's vocabulary loss (0.40) is the smallest of any degradation type, closely matching DeepSeek R1's pattern (0.41). Both models rank vocabulary simplification as the least impactful manipulation. This cross-model convergence strengthens the finding that vocabulary is a secondary quality signal for LLM evaluators.

\textbf{Both models agree on the broad qualitative pattern}: vocabulary remains the weakest manipulation for both models, while voice, structure, and combined degradation produce the largest losses. In the available QwQ degradation outputs, voice loss is larger than in DeepSeek (4.50 vs.\ 2.34), suggesting greater sensitivity to authorial distinctiveness. Because this comparison is based on a single QwQ degradation run with one failed Fitzgerald voice evaluation excluded, we treat it as directional rather than definitive.


\section{Discussion}

\subsection{Connecting the Two Studies}

Study 1 extracted the model's stated theory: quality = craft + depth + voice, with craft weighted most heavily. Study 2 probed the model's \emph{operative} sensitivity to these features and found partial alignment:

\textbf{Confirmed}: Craft (operationalized as sentence structure) is the feature whose degradation most reduces the model's quality scores. Structure degradation produces the largest single-feature score reduction.

\textbf{Refined}: Voice matters, but \emph{author-specifically}. The model's stated theory treats voice as a general quality dimension; the sensitivity analysis reveals it as author-dependent in the model's evaluations. Woolf's quality is voice-dependent; Orwell's is not.

\textbf{Challenged}: The model's reasoning traces frequently praise ``precise vocabulary'' and ``word choice'' in high-quality texts. Yet vocabulary simplification produces only minimal score reduction ($0.41 \pm 0.46$), an order of magnitude smaller than structure or voice effects. The model \emph{describes} vocabulary as important but its evaluations are \emph{relatively insensitive} to vocabulary changes. This gap between stated and revealed preferences---consistent with findings on reasoning faithfulness \citep{turpin2024language,lanham2023measuring}---is a significant finding.

\subsection{The Vocabulary Paradox}

Vocabulary simplification produces the smallest quality loss of any degradation type ($0.41 \pm 0.46$ across replications)---an order of magnitude smaller than structure ($2.8$) or voice ($2.3$) loss. This relative insensitivity contrasts sharply with the model's reasoning traces, which frequently praise ``precise vocabulary'' and ``word choice'' in high-quality texts. Several explanations are possible:

\textbf{Structure dominates vocabulary as a quality signal.} When vocabulary is simplified but structure preserved, the architectural sophistication may signal quality regardless of word choice. The model appears to attend to structure so strongly that vocabulary becomes a secondary signal.

\textbf{Canonical recognition partially buffers vocabulary changes.} As documented in Section~\ref{sec:recognition}, the model explicitly identifies 4 of 5 literary texts. Recognition could buffer vocabulary-simplified scores if the model ``knows'' it is reading Morrison and anchors accordingly. However, this explanation is weakened by the fact that combined degradation \emph{does} reduce scores dramatically---the model isn't simply pattern-matching to known passages. The selective nature of the effect (minimal vocabulary loss but substantial structure loss) also argues against pure reputation scoring.

\textbf{Vocabulary is genuinely secondary for this evaluator.} Great writers from Hemingway to Orwell have used plain vocabulary to powerful effect. The model's \emph{operative} sensitivity (as revealed by degradation) diverges from its \emph{stated} sensitivity (as articulated in reasoning traces)---a finding consistent with research on reasoning faithfulness \citep{turpin2024language}.

\subsection{Combined Degradation and Feature Interaction}

A surprising finding from replications is that combined degradation ($-5.6 \pm 1.9$) is \emph{subadditive}: it produces a smaller loss than the sum of individual degradations ($8.1 \pm 4.8$; Wilcoxon $p = 0.99$ for superadditivity). This reverses our initial single-run observation and illustrates the importance of replication. The initial margin of $+0.10$ was noise.

The subadditivity likely reflects a quality floor effect: individual degradations that each reduce scores substantially cannot compound beyond the minimum of the scale. When structure loss already brings a text to 4/10, adding vocabulary loss cannot push it much lower. This is a ceiling/floor artifact rather than evidence about feature interaction.

Nevertheless, combined degradation consistently produces the largest absolute score reductions ($5.6$ points), confirming that stripping all literary features simultaneously is devastating. The practical implication for AI text generation is that maintaining quality requires attention to multiple features simultaneously, even if the worst case is bounded by a quality floor.

\subsection{Implications for Writing Instruction}

Our findings suggest that LLM quality assessment is more sensitive to structural features than to vocabulary, which has potential implications for how automated writing feedback is designed:

\begin{itemize}
\item \textbf{Structure over vocabulary.} In the model's evaluations, sentence architecture is the feature whose degradation most reduces quality scores, while vocabulary sophistication has a much smaller effect. If LLMs are used for automated writing feedback, this sensitivity pattern would systematically underweight vocabulary instruction.
\item \textbf{Voice is author-specific.} Different writers' scores respond to different degradation types. This suggests that automated feedback systems should avoid prescribing universal rules about ``voice.''
\item \textbf{Feature interaction matters.} The large combined degradation effect suggests that quality features interact in the model's assessments. Automated feedback that addresses features independently may miss these interactions.
\end{itemize}

We emphasize that these are observations about the \emph{model's} sensitivity patterns, not about writing quality per se. Whether human readers show similar sensitivity patterns is an open empirical question.

\subsection{Limitations}

Several limitations constrain our conclusions:

\begin{itemize}
\item \textbf{Reasoning traces are not transparent windows into computation}: A fundamental caveat is that reasoning traces reflect the model's \emph{stated} rationale, not necessarily its internal decision process. \citet{turpin2024language} show that chain-of-thought explanations can be systematically unfaithful, and \citet{lanham2023measuring} find that stated reasoning often diverges from operative features. The vocabulary paradox---where the model praises vocabulary but shows no sensitivity to its degradation---may be an instance of this disconnect. We treat reasoning traces as suggestive data about the model's processing, not as ground truth about its computations.

\item \textbf{Training data contamination and source recognition}: As shown in Section~\ref{sec:recognition}, the model explicitly identifies 6 of 30 texts by author or title in its reasoning traces, concentrated entirely in the top two tiers. Our familiarity experiment finds suggestive evidence that recognition contributes to score inflation, but this is confounded by the genuine quality gap between canonical prose and researcher-written pastiches. This confound is particularly acute for the vocabulary finding: the model's insensitivity to vocabulary simplification could partly reflect recognition-based anchoring (the model ``knows'' it is reading Morrison regardless of word choice). Future work should test with high-quality texts by accomplished but non-canonical authors.

\item \textbf{Cross-model validation}: We report results from both DeepSeek R1 and Qwen QwQ (Section~\ref{sec:cross_model}), but both are reasoning-enabled LLMs trained on overlapping internet corpora. Models with fundamentally different architectures or training data may show different sensitivity patterns.

\item \textbf{Manual degradation}: Our degradations were researcher-created; different researchers might make different choices. Algorithmic degradation would improve reproducibility.

\item \textbf{Sample size}: Five texts for the degradation study, with five replications each, provides moderate statistical power. We report confidence intervals throughout and note where effects do not reach significance.

\item \textbf{Prompt sensitivity}: We use a single evaluation prompt throughout. Different prompts could elicit different sensitivity patterns, and prompt sensitivity remains untested.

\item \textbf{No human baseline}: We lack systematic human judgments for comparison. A human study would contextualize whether the model's sensitivity patterns match human perception.
\end{itemize}

\section{Conclusion}

We have presented a study that both extracts and empirically probes an LLM's implicit theory of literary quality. Through observation (Study 1), we identified that the model's stated theory values intentional craft, layered depth, and distinctive voice. Through sensitivity analysis (Study 2), replicated across five independent DeepSeek runs and compared directionally with a second model, we found:

\begin{enumerate}
\item \textbf{Vocabulary simplification produces only a small reduction in the model's quality scores.} The model's assessments are far more sensitive to structure than to vocabulary, though source recognition may partially confound this finding.
\item \textbf{Structure is the most impactful single feature.} Sentence architecture degradation produces the largest score reductions across both models and all replications.
\item \textbf{Sensitivity patterns are author-specific.} Different texts show different vulnerability profiles: Woolf's scores are voice-dependent; Orwell's are not. There is no universal sensitivity formula.
\item \textbf{Combined degradation is devastating but subadditive.} Stripping all literary features simultaneously produces the largest score reductions ($-5.6$ points), but this is less than the sum of individual losses, likely due to a quality floor effect. The initial single-run observation of superadditivity did not replicate.
\end{enumerate}

These findings contribute to computational aesthetics, inform the design of automated writing feedback systems, and provide methodological tools---degradation-based sensitivity analysis with replication and cross-model validation---for investigating LLM evaluation behavior. The question ``What is good?'' remains open, but we now have empirical tools to investigate how LLMs answer it.

\begingroup
\small
\ifanonymized\else
\paragraph{Acknowledgments.}

This work was conducted at Uppsala University. We thank the UppsalaNLP group for discussions on text quality and evaluation methodology.
\fi

\paragraph{Data Availability Statement.}

A reproducibility package containing the prompts, model outputs, replication files, analysis scripts, and derived benchmark metadata for this study is available in the project \href{https://github.com/BirgerMoell/WhatIsGood}{GitHub repository}. Because some source passages are drawn from copyrighted works, the public repository provides metadata, derived annotations, and reconstruction guidance rather than full source excerpts in every case.
\endgroup

\bibliographystyle{plainnat}
\bibliography{references}

\begin{thebibliography}{18}
\providecommand{\natexlab}[1]{#1}
\providecommand{\url}[1]{\texttt{#1}}
\expandafter\ifx\csname urlstyle\endcsname\relax
  \providecommand{\doi}[1]{doi: #1}\else
  \providecommand{\doi}{doi: \begingroup \urlstyle{rm}\Url}\fi

\bibitem[Aristotle(-350)]{aristotle_rhetoric}
Aristotle.
\newblock \emph{Rhetoric}.
\newblock Dover Publications, -350.
\newblock Translated by W. Rhys Roberts, 1924.

\bibitem[Brooks(1947)]{brooks1947well}
Cleanth Brooks.
\newblock \emph{The Well Wrought Urn: Studies in the Structure of Poetry}.
\newblock Harcourt Brace, 1947.

\bibitem[Elbow(1981)]{elbow1981writing}
Peter Elbow.
\newblock \emph{Writing with Power: Techniques for Mastering the Writing
  Process}.
\newblock Oxford University Press, 1981.

\bibitem[Gardner(1991)]{gardner1991art}
John Gardner.
\newblock \emph{The Art of Fiction: Notes on Craft for Young Writers}.
\newblock Vintage Books, 1991.

\bibitem[Gunning(1952)]{gunning1952technique}
Robert Gunning.
\newblock The technique of clear writing.
\newblock \emph{McGraw-Hill}, 1952.

\bibitem[Kao and Jurafsky(2012)]{kao2012computational}
Justine Kao and Dan Jurafsky.
\newblock A computational analysis of style, affect, and imagery in
  contemporary poetry.
\newblock In \emph{Proceedings of the NAACL-HLT Workshop on Computational
  Linguistics for Literature}, pages 8--17, 2012.

\bibitem[Ke and Ng(2019)]{ke2019automated}
Zixuan Ke and Vincent Ng.
\newblock Automated essay scoring: A survey of the state of the art.
\newblock In \emph{Proceedings of the International Joint Conference on
  Artificial Intelligence (IJCAI)}, pages 6300--6308, 2019.

\bibitem[Kincaid et~al.(1975)Kincaid, Fishburne~Jr, Rogers, and
  Chissom]{kincaid1975derivation}
J.~Peter Kincaid, Robert~P. Fishburne~Jr, Richard~L. Rogers, and Brad~S.
  Chissom.
\newblock Derivation of new readability formulas for navy enlisted personnel.
\newblock \emph{Research Branch Report}, 8\penalty0 (75), 1975.

\bibitem[Koo et~al.(2024)Koo, Lee, Raheja, Park, Kim, and
  Kang]{koo2024benchmarking}
Ryan Koo, Minhwa Lee, Vipul Raheja, Jong~Inn Park, Zae~Myung Kim, and Dongyeop
  Kang.
\newblock Benchmarking cognitive biases in large language models as evaluators.
\newblock \emph{arXiv preprint arXiv:2309.17012}, 2024.

\bibitem[Lanham et~al.(2023)Lanham, Chen, Radhakrishnan, Steiner, Denison,
  Hernandez, Li, Durmus, Hubinger, Kernion, et~al.]{lanham2023measuring}
Tamera Lanham, Anna Chen, Ansh Radhakrishnan, Benoit Steiner, Carson Denison,
  Danny Hernandez, Dustin Li, Esin Durmus, Evan Hubinger, Jackson Kernion,
  et~al.
\newblock Measuring faithfulness in chain-of-thought reasoning.
\newblock \emph{arXiv preprint arXiv:2307.13702}, 2023.

\bibitem[Lee et~al.(2024)Lee, Phatale, Mansoor, Lu, Mesnard, Bishop, Carbune,
  and Rastogi]{lee2024rlaif}
Harrison Lee, Samrat Phatale, Hassan Mansoor, Kellie Lu, Thomas Mesnard, Colton
  Bishop, Victor Carbune, and Abhinav Rastogi.
\newblock Rlaif: Scaling reinforcement learning from human feedback with {AI}
  feedback.
\newblock \emph{arXiv preprint arXiv:2309.00267}, 2024.

\bibitem[Liu et~al.(2023)Liu, Iter, Xu, Wang, Xu, and Zhu]{liu2023geval}
Yang Liu, Dan Iter, Yichong Xu, Shuohang Wang, Ruochen Xu, and Chenguang Zhu.
\newblock G-{E}val: {NLG} evaluation using {GPT-4} with better human alignment.
\newblock In \emph{Proceedings of EMNLP}, 2023.

\bibitem[Stamatatos(2009)]{stamatatos2009survey}
Efstathios Stamatatos.
\newblock A survey of modern authorship attribution methods.
\newblock \emph{Journal of the American Society for Information Science and
  Technology}, 60\penalty0 (3):\penalty0 538--556, 2009.

\bibitem[Turpin et~al.(2024)Turpin, Michael, Perez, and
  Bowman]{turpin2024language}
Miles Turpin, Julian Michael, Ethan Perez, and Samuel~R. Bowman.
\newblock Language models don't always say what they think: Unfaithful
  explanations in chain-of-thought prompting.
\newblock In \emph{Advances in Neural Information Processing Systems},
  volume~36, 2024.

\bibitem[Uto(2021)]{uto2021review}
Masaki Uto.
\newblock A review of deep-neural automated essay scoring models.
\newblock \emph{Behaviormetrika}, 48:\penalty0 459--484, 2021.

\bibitem[Wang et~al.(2024)Wang, Li, Chen, Cai, Zhu, Lin, Cao, Kong, Liu, Liu,
  and Sui]{wang2024large}
Peiyi Wang, Lei Li, Liang Chen, Feifan Cai, Dawei Zhu, Binghuai Lin, Yunbo Cao,
  Lingpeng Kong, Qi~Liu, Tianyu Liu, and Zhifang Sui.
\newblock Large language models are not fair evaluators.
\newblock \emph{arXiv preprint arXiv:2305.17926}, 2024.

\bibitem[Wilson and Schooler(1991)]{wilson1991thinking}
Timothy~D. Wilson and Jonathan~W. Schooler.
\newblock Thinking too much: Introspection can reduce the quality of
  preferences and decisions.
\newblock \emph{Journal of Personality and Social Psychology}, 60\penalty0
  (2):\penalty0 181--192, 1991.

\bibitem[Zheng et~al.(2023)Zheng, Chiang, Sheng, Zhuang, Wu, Zhuang, Lin, Li,
  Li, Xing, et~al.]{zheng2023judging}
Lianmin Zheng, Wei-Lin Chiang, Ying Sheng, Siyuan Zhuang, Zhanghao Wu, Yonghao
  Zhuang, Zi~Lin, Zhuohan Li, Dacheng Li, Eric Xing, et~al.
\newblock Judging {LLM}-as-a-judge with {MT-Bench} and chatbot arena.
\newblock \emph{Advances in Neural Information Processing Systems}, 36, 2023.

\end{thebibliography}

\end{document}